\pdfoutput=1
\documentclass[a4paper]{article}
\usepackage{spconf,amsmath,graphicx}
\usepackage{multirow}
\usepackage{makecell}
\usepackage{multicol}
\usepackage{tikz}
\usepackage{hyperref}
\usetikzlibrary{arrows,positioning} 
\usetikzlibrary{shapes,decorations}
\usetikzlibrary{calc}
\usetikzlibrary{decorations.pathreplacing}
\usepackage{xcolor}

\definecolor{blind_red}{HTML}{D7191C}
\definecolor{blind_orange}{HTML}{FDAE61}
\definecolor{blind_yellow}{HTML}{FFFFBF}
\definecolor{blind_blue}{HTML}{ABD9E9}
\definecolor{blind_blue2}{HTML}{2C7BB6}
\def\x{{\mathbf x}}

\copyrightnotice{979-8-3503-0689-7/23/\$31.00~\copyright2023 IEEE}

\newcommand\blfootnote[1]{%
  \begingroup
  \renewcommand\thefootnote{}\footnote{#1}%
  \addtocounter{footnote}{-1}%
  \endgroup
}

\title{On the Relevance of Phoneme Duration Variability of Synthesized Training Data for Automatic Speech Recognition}

\name{Nick Rossenbach$^{\dagger,*}$, Benedikt Hilmes$^*$, Ralf Schlüter$^\dagger$}
\address{Machine Learning and Human Language Technology,\\Computer Science Departement, RWTH Aachen University, Germany\\$^\dagger$AppTek GmbH, Germany\\\texttt{\{rossenbach, hilmes, schlueter\}@cs.rwth-aachen.de} }
\newcommand{\alert}[1]{#1}

\begin{document}
%
\maketitle
\begin{abstract}
Synthetic data generated by text-to-speech (TTS) systems can be used to improve automatic speech recognition (ASR) systems in low-resource or domain mismatch tasks.
It has been shown that TTS-generated outputs still do not have the same qualities as real data.
In this work we focus on the temporal structure of synthetic data and its relation to ASR training.
By using a novel oracle setup we show how much the degradation of synthetic data quality is influenced by duration modeling in non-autoregressive (NAR) TTS. 
To get reference phoneme durations we \alert{use} two common alignment methods, a hidden Markov Gaussian-mixture model (HMM-GMM) aligner and a neural connectionist temporal classification (CTC) aligner.
\alert{Using a simple algorithm based on random walks we shift phoneme duration distributions of the TTS system closer to real durations, resulting in an improvement of an ASR system using synthetic data in a semi-supervised setting.}
\end{abstract}
\blfootnote{\vspace{-1em}$^*$\textit{\small equal contribution}}
\section{Introduction}
Previous literature showed that it is possible to improve automatic speech recognition (ASR) systems using synthetic data generated via text-to-speech (TTS) \cite {Baskar-2019-Semi-SupervisedSequ,Rosenberg-2019-SpeechRecognitionw,Rossenbach-2020-GeneratingSynthetic,Laptev-2020-YouDoNotNeedMore,Baskar-2021-EatEnhancedASR-TT}.
Besides directly improving a single recognition task with synthetic data from additional text, different kinds of domain adaptation are possible via synthetic data \cite{Rosenberg-2019-SpeechRecognitionw,Zheng-2021-UsingSyntheticAudi}.
With a few exceptions \cite{9688255}, only ``end-to-end'' neural speech recognition systems were used in such works. The most notable architectures are the transducers \cite{8268937} and the attention-encoder-decoder \cite{7472621}. For both architectures it has been shown that adding synthetic data via text-to-speech systems can improve the training or adaptation of the recognition models. For other architectures this seems to be more problematic \cite{9688255}. \alert{Most prior work to create synthetic data has been done by using autoregressive TTS systems, most of them similar to Tacotron-2 \cite{Shen-2018-NaturalTTSSynthesi}, with \cite{9688218} as notable exception.}
Currently, literature shows a trend towards non-autoregressive TTS \cite{Ren-2020-FastSpeech2Fasta, 9413889, pmlr-v139-kim21f} as it is claimed to be more robust and controllable. \alert{Non-autoregressive TTS systems usually use an explicit prediction of phoneme durations, and in many cases also explicit training targets from an external aligner. This allows for new experimental possibilities to investigate the influence of temporal control within the scope of synthetic training data creation via TTS.} 

\vspace{-0.4em}
\subsection{Related Work}
\vspace{-0.2em}
Within the scope of improving the use of synthetic data different approaches were published to deal with the mismatch of synthetic data compared to real data.
\cite{Baskar-2021-EatEnhancedASR-TT} proposed to reduce the influence of the synthetic data with respect to acoustic modeling by weighting the acoustic encoder context lower for synthetic inputs.
\cite{Hu-2022-SYNTUtilizingIm} introduced rejection sampling to exclude non-matching synthetic examples as well as separate statistics for the batch norm layers for real and synthetic data.
\cite{Zheng-2021-UsingSyntheticAudi} proposed an elastic weight penalty between the model weights trained on real data and the current model weights.
This limits the model degradation when shifting domain.
While many papers study the effect of synthetic data on ASR systems or the performance of TTS in general, there is a lack of analysis on the difference between real and synthetic audio data.
We only know of \cite{https://doi.org/10.48550/arxiv.2211.16049} investigating this issue.
\alert{Additionally, w.r.t. alignment methods for TTS, previous work showed that the influence of the chosen method is rather limited \cite{Zalkow-2023-EvaluatingSpeechPh}.}

\vspace{-0.4em}
\subsection{Contribution}
\vspace{-0.3em}
In contrast to changing the ASR model to make better use of synthetic training data, we investigate what synthetic data is currently lacking. In specific, we want to have a look at the temporal diversity of synthetic data and its implications on the training process. While it may be obvious that synthetic data has deficiencies in temporal diversity, we want to showcase how much this affects the ASR training process. Different to other publications, we will not focus on general feature or speaker representation mismatches in the scope of this work, but only on the temporal aspect. To obtain reference alignments, we \alert{use} two different common aligner approaches for the TTS training, an ``HMM-GMM'' (c.f. \cite{McAuliffe-2017-MontrealForcedAlig}) and a ``neural CTC aligner'' approach (c.f. \cite{Perez-Gonzalez-de-Martos-2021-VRAIN-UPVMLLPssys}). \alert{We use two different aligners to show that our findings are independent of the chosen aligning method.}
Both aligners are commonly used in NAR-TTS literature, and we show that in our case using the HMM-GMM results in slightly better performance, although the CTC aligner has similar potential in an oracle ``duration cheating'' setup. We show that generally the TTS model is under-predicting phoneme durations in their mean and variance compared to the given alignment and analyze methods to mitigate this mismatch. Finally, we are proposing to modify the internal duration prediction using a random-walk-based algorithm to increase the temporal variation of synthetic data without any change to the TTS training itself.
In a semi-supervised training setting on the English LibriSpeech \cite{librispeech} task we show for the first time that also NAR-TTS systems can be capable of improving an ASR system with synthetic data, achieving even better results than currently published in literature \cite{Baskar-2021-EatEnhancedASR-TT,9688255}. These results are then further improved by using the proposed random-walk modification of the TTS.
We follow an open code approach, and our training recipes were publicly accessible during the creation of this work\footnote{\url{https://github.com/rwth-i6/returnn-experiments/tree/master/2023-phoneme-duration-variability}}. As toolkits we used RASR \cite{Wiesler-2014-RASRNNTheRWTHne}, RETURNN \cite{DBLP:conf/icassp/DoetschZVKSN17} and the workflow manager Sisyphus \cite{DBLP:conf/emnlp/PeterBN18}.
\vspace{-0.9em}
\section{Speech Synthesis}
\vspace{-0.4em}
\subsection{Feature prediction model}
\label{sec:nar_tts}
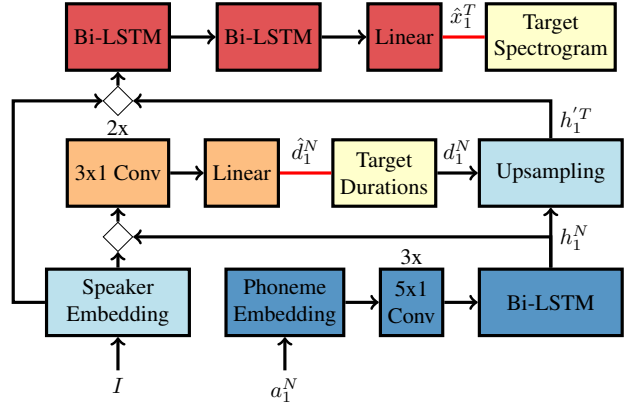
\begin{figure}
	\begin{center}
	\resizebox{3.2in}{!}{
	\begin{tikzpicture}
		\tikzstyle{Block} = [rectangle,
thick,
line width=0.5mm,
minimum height = 1em,
draw=black,
text centered,
minimum height=3em]
	\tikzstyle{Block2} = [rectangle,
thick,
line width=0.5mm,
minimum height = 1em,
draw=black,
text centered,
minimum height=3em]
			\node[Block2, fill=blind_blue2!80, align=center] (emb) at (0, 0) {Phoneme\\ Embedding};
			\node[Block2, fill=blind_blue2!80, right =15pt of emb, align=center] (conv) {5x1 \\ Conv};
			\node (f) [below = 15 pt of emb] {$a_1^N$};
			\node (convtext) [above = -2 pt of conv] {3x};
			\node (enclstm) [Block2, fill=blind_blue2!80, right = 15pt of conv, minimum width=2.25cm] {Bi-LSTM};
			\node (speakemb) [Block2, fill=blind_blue!80, left = 15 pt of emb, align=center, minimum width=2.25cm] {Speaker\\ Embedding};
			\node (token) [below = 15 pt of speakemb] {$I$};
			
			\node (durconv) [Block2, fill=blind_orange!80, above = 27 pt of speakemb, align = center, minimum width=1.5cm] {3x1 Conv};
			\node (convtext2) [above = -2 pt of durconv] {2x};
			\node (durlin) [Block2, fill=blind_orange!80, right= 15 pt of durconv] {Linear};
			\node (durtarget) [Block2, fill=blind_yellow!80, right=23 pt of durlin, align=center] {Target\\ Durations};
			\node (durcat) [diamond, above=7pt of speakemb, draw]{};
			\node (upsampling) [Block2, fill=blind_blue!80, above=27pt of enclstm, align=center, minimum width=2.25cm] {Upsampling};
			
			\node (declstm1) [Block, fill=blind_red!75, above = 30 pt of durconv] {Bi-LSTM};
			\node (declstm2) [Block, fill=blind_red!75, right = 20pt of declstm1] {Bi-LSTM};
			\node (declin) [Block, fill=blind_red!75, right = 20pt of declstm2] {Linear};
			\node (dectarget) [Block, fill=blind_yellow!80, above=30 pt of upsampling, align=center] {Target \\ Spectrogram};
			\node (upcat) [diamond, above=10pt of durconv, draw]{};
			
			\draw [->, line width=0.5mm] (f) -- (emb);
			\draw [->, line width=0.5mm] (token) -- (speakemb);
			\draw [->, line width=0.5mm] (emb) -- (conv);
			\draw [->, line width=0.5mm] (conv) -- (enclstm);

			\draw let \p1 = (durcat), \p2 = (enclstm) in [->,smooth, line width=0.5mm] ([shift={(0,0)}]enclstm.north)
			-- (\x2,\y1)
			-- (durcat.east);
			\draw [->, line width=0.5mm] (speakemb.north) -- (durcat);
			\draw [->, line width=0.5mm] (durcat) -- (durconv);
			\draw [->, line width=0.5mm] (durconv) -- (durlin);
			\draw [line width=0.5mm, red] (durlin) -- (durtarget) node[midway, above, black] {$\hat{d}_1^N$};
			\draw [->, line width=0.5mm] (enclstm) -- (upsampling) node[midway, right] {$h_1^N$};
			\draw [->, line width=0.5mm] (durtarget) -- (upsampling) node[midway, above] {$d_1^N$};
			
			\draw let \p1 = (upcat), \p2 = ([shift={(-0.5,0)}]speakemb.west) in [->, line width=0.5mm] (speakemb.west) -- ([shift={(-0.5, 0)}]speakemb.west) -- (\x2, \y1) -- (upcat.west);
			\draw let \p1 = (upcat), \p2 = (upsampling) in [->, line width=0.5mm] (upsampling) -- node[midway, right, black] {$h^{'T}_1$} (\x2, \y1) -- (upcat);
			\draw [->, line width=0.5mm] (upcat) -- (declstm1);
			\draw [->, line width=0.5mm] (declstm1) -- (declstm2);
			\draw [->, line width=0.5mm] (declstm2) -- (declin);
			\draw [line width=0.5mm, red] (declin) -- (dectarget) node[midway, above, black] {$\hat{x}_1^T$};
		\end{tikzpicture}}
	\end{center}
		\vspace{-1em}
	\caption{Non-autoregressive TTS model. The encoder part is depicted in blue, the duration predictor in orange and the decoder in red. The red lines mark training losses.}
	\label{fig:nartts}
	\vspace{-1em}
\end{figure}

Our TTS system does not directly follow any previous publication, but is related to the model presented in \cite{Perez-Gonzalez-de-Martos-2021-VRAIN-UPVMLLPssys} with Gaussian upsampling \cite{Shen-2020-Non-AttentiveTacotr}. The architecture is depicted in Figure \ref{fig:nartts}.
The phoneme encoder consists of three 256-dimensional convolutional layers with filter size 5, ReLU activation and batch-normalization, followed by one bi-directional LSTM (BLSTM) layer \cite{hochreiter1997lstm} with 256 dimensions per direction.
The duration predictor consists of two 256-dimensional convolutions with filter size 3, ReLU activation and layer normalization.
The duration prediction itself is modeled by a linear layer with a scalar softplus output.
The decoder consists of two BLSTM layers with 1024 hidden states per direction and a linear layer for spectrogram prediction. During training the target durations are used for upsampling. We use globally normalized 80-dimensional log-mel features with frame shift 12.5 ms and window size of 50 ms. \alert{The loss for $N$ phonemes and $T$ audio frames is defined as:
\vspace{-0.3em}
\begin{align}
L_p &= \sum_{n=1}^N |\hat{d}_n - d_n| \\
L_f &= \frac{1}{80} \sum_{t=1}^T ||\hat{x}_t - x_t||_1 \\
L &= L_p + L_f
\end{align} 
with target features $x_t$, target durations $d_n$, and $\hat{x}_t$ and $\hat{d}_n$ being the respective predictions.}
Multi-speaker capabilities are enabled by passing a speaker ID to a trainable look-up table.
The speaker information is added as input to the decoder and duration predictor.
During prediction, the log-mel features are converted into 512 dimensional linear features needed for Griffin\&Lim (G\&L) \cite{DBLP:conf/icassp/GriffinDL84} vocoding using a separately trained 2x1024-dimensional BLSTM layer network, resulting in a total of 63M parameters.
The phoneme set consists of ARPABET phoneme symbols without stress marker. We insert a \texttt{[space]} token between the phonemes of words to mark both a word boundary and possible silence. We use Sequitur \cite{Bisani-2008-Joint-sequencemodel} to predict phoneme sequences for words not part of the LibriSpeech lexicon.

\subsection{HMM-GMM Aligner}
\label{sec:hmm_gmm}

Our HMM-GMM aligner implemented in RASR is similar to the Montreal-Forced-Aligner (MFA) \cite{McAuliffe-2017-MontrealForcedAlig} used in TTS publications such as FastSpeech-2 \cite{Ren-2020-FastSpeech2Fasta}. Its parameters have been optimized on the LibriSpeech-100h ASR task \cite{librispeech} and includes the following training steps:

\alert{
\begin{enumerate}
\setlength\itemsep{0.0em}
\item Feature extraction and initial alignment: We use 16-dimensional MFCC features with an additional energy component to discriminate silence and non-silence frames. Then, we distribute the non-silence frames equally over the sequence for an initial alignment.
\item Monophone training: We perform 75 iterations of expectation-maximization (EM) algorithm training using single density Gaussian mixtures over the features and 10 iterations of splitting and re-estimating the then multi-modal mixtures.
\item State-tying: We use classification and regression trees (CART) \cite{young1992cart} to create 12k HMM state labels covering triphone clusters.
\item Triphone training: We apply a context window of 9 on the MFCC features and use linear discriminant analysis (LDA) to reduce the feature size from 144 to 48. Based on the alignment from step 2, EM training is performed including mixture splits for 10 iterations. 
\item Speaker-Adaptation: Based on the speaker labels, a fMLLR transformation is learned on the audio features of non-silence frames \cite{Gales-1998-Maximumlikelihoodl}. Afterwards the mixtures are re-estimated on the resulting features with applied linear transformation.
\end{enumerate}
}
The model uses three HMM states per phoneme and one for silence.
The duration is extracted by using the Viterbi algorithm to find the best alignment.
As for HMMs silence between words is optional, we assign a duration of zero to \texttt{[space]} tokens that do not have corresponding silence in the alignment.

\subsection{CTC Aligner}

The neural alignment approach is a frame-wise label prediction neural network using CTC loss. The model architecture is designed following \cite{Perez-Gonzalez-de-Martos-2021-VRAIN-UPVMLLPssys}. The encoder consists of 5 256-dimensional convolutional layers with filter size 5, one 512-dimensional BLSTM layer and a linear projection with softmax activation as encoder.
The reconstruction network consists of two 512-dimensional BLSTM layers and a linear projection as output, resulting in a total of 15M parameters. The input to the reconstruction is the probability tensor of the encoder softmax. \alert{The target features are the same log-mel features as for the TTS system defined in Section \ref{sec:nar_tts}.
The target loss $L$ is defined as:
\begin{equation}
L_{ctc} = CTC\left(p_{enc}({y'}_1^T | x_1^T, s), y_1^N\right)
\end{equation}
\begin{equation}
L_{rec} = \frac{1}{80} \sum_{t=1}^T||g_{rec}\left(p_{enc}({y'}_t | x_1^T, s), s\right) - x_t||_2^2
\end{equation}
\begin{equation}
L = L_{ctc} + \lambda L_{rec}
\end{equation}
for features $x_1^T$, transcription token sequence $y_1^N$, frame-wise label token sequence including blank ${y'}_1^T$ and speaker index~$s$. $CTC$ denotes the loss value resulting from the Baum-Welch algorithm using the CTC topology. $p_{enc}$ denotes the probability estimation via the encoder and $g_{rec}$ the feature reconstruction of the reconstruction model. In all experiments we used $\lambda = 0.5$.}
The CTC aligner is fully speaker dependent, as we are concatenating the speaker information via a lookup table to the input at the first encoder and reconstruction layer. Feeding the speaker label to the encoder is not necessary, but resulted in faster alignment convergence.
For the Viterbi alignment we set the probability of the blank symbol to a small non-zero number, so that we have non-blank labels at each frame position.
Note that in the case of CTC \texttt{[space]} tokens are treated as regular phonemes and get a minimum duration of 1. This is different to the alignments from the HMM-GMM system, where silence is modeled separately from the phonemes and only optionally inserted between words.
\vspace{-0.3em}
\section{Automatic Speech Recognition}
\vspace{-0.2em}
We use an attention-encoder-decoder (AED) ASR system \cite{7472621, Zeyer-2018-ImprovedTrainingof} to evaluate training on synthetic data and mixed data.
For the neural encoder model we follow the Conformer architecture \cite{gulati20_interspeech}.
The model consists of 12 512-dimensional Conformer layers in the encoder part and a single 1024-dimensional LSTM as decoder and has a total of 98M parameters.
We use 80-dimensional log-mel features with a window size of 25ms and 10ms shift. We downsample the features with 2 BLSTM-layers and max-pooling with factors 3 and 2, resulting in a reduction factor of 6.
In order to stabilize the training we gradually increase the number of encoder layers in the beginning of the training, starting with 2 layers and adding 2 layers each 5 sub-epochs steps.
We apply SpecAugment \cite{park19e_interspeech} and speed-perturbation via librosa.resample()\footnote{\url{https://librosa.org/doc/latest/generated/librosa.resample.html}} with uniformly distributed scaling factors 0.9/1.0/1.1 on the input audio. For some experiments, we use a 24-layer Transformer language model (305M parameters) with shallow fusion during recognition \cite{irie19b_interspeech}.

\vspace{-0.3em}
\section{Pipeline and Method}
\vspace{-0.2em}
\begin{figure}
	\begin{center}
	\resizebox{3.2in}{!}{
	\begin{tikzpicture}[auto]
		\node[rectangle, minimum size=1.75cm, draw, fill=blind_blue!80, align=center] (gmm) at (0, 0) {Aligner \\ training};
		\node[rectangle, minimum size=1.75cm,draw, fill=blind_blue!80, right=20 pt of gmm, align=center] (align) {Forced \\ align};
		\node[rectangle, minimum size=1.75cm,draw,  fill=blind_orange!80, right=20 pt of align, align=center] (tts) {TTS \\ training};
		\node[rectangle, minimum size=1.75cm,draw,  fill=blind_orange!80, right=20 pt of tts, align=center] (synth) {Synthesis};
		\node[rectangle, minimum size=1.75cm,draw,  fill=blind_red!75, right=20 pt of synth,align=center] (asr) {ASR \\ training};
		\node[below= 18 pt of synth,align=center] (real) {random \\ speaker tag};
		\node[above= 18 pt of gmm] (data) {preprocessed data};
		\node[above= 18 pt of asr] (realdata) {original data};
		\draw [->, line width=0.5mm] (gmm) -- (align);
		\draw [->, line width=0.5mm] (align) -- (tts);
		\draw [->, line width=0.5mm] (tts) -- (synth);
		\draw [->, line width=0.5mm] (synth) -- (asr);
		\path (align.south east) edge[bend right,->, line width=0.5mm, dashed] node [below=4pt, align=center] {oracle \\ durations} (synth.south west);
		\draw [->, dashed, line width=0.5mm] (real) -- (synth);
		\draw [->, line width=0.5mm] (data) -- (gmm);
		\path (data.east) edge[->, bend left = 18, line width=0.3mm] (tts.north west);
		\draw [->, dashed, line width=0.5mm] (realdata) -- (asr);
		
	\end{tikzpicture}
	}
	\end{center}
	\vspace{-1.5em}
	\caption[Experiment pipeline]{Experiment pipeline for synthetic data training.}
	\label{fig:pipeline}
	\vspace{-0.5em}
\end{figure}
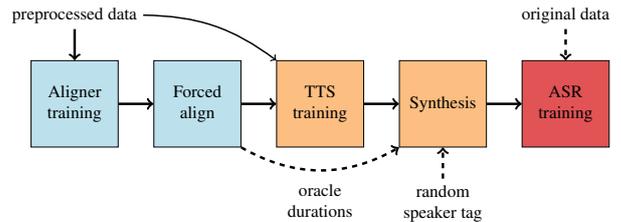

The experimental pipeline consists of pre-processing and 5 main steps, visualized in Figure \ref{fig:pipeline}. First, excessive silence is stripped from the data using the silence filter from FFmpeg\footnote{\url{https://ffmpeg.org/ffmpeg-filters.html\#silenceremove}} with a threshold of -50dB. This data is used for both the aligner and TTS training. In order to determine the phoneme durations for TTS training we use the aligner on the full pre-processed data, including the cross validation data. \alert{After the TTS training we synthesize either the training text or unseen text data. As the TTS system is trained with fixed speaker labels we are randomizing the speaker IDs during synthesis. The synthesized data is used to train the final ASR model. In the first case we are only using the synthesized data on the training corpus itself to train the ASR model. We do this in order to investigate the gap between the real and synthetic data using exactly the same text information. In a second setting, we use additional text to create additional synthetic data, and train the ASR system on a combination of both real and synthetic data.}

\subsection{Oracle Durations}

In order to evaluate the degradation of the synthetic data with respect to phoneme durations, we synthesize the data using the durations extracted directly from the original forced alignment using either the HMM-GMM or the CTC aligner. By doing this we have two goals in mind:
\begin{enumerate}
\item We can directly measure in which way the predictions from the duration predictor are deviating from the ground truth.
\item We can compare how the choice of the alignment system influences the synthetic data generation.
\end{enumerate}
\alert{By training an ASR system on the synthetic data generated from using the phoneme durations directly from the forced alignment, we can estimate an upper bound for the performance gain from having a more realistic temporal diversity. This makes it possible to see how close the predicted phoneme durations perform to the underlying real phoneme durations.}

\subsection{Duration Modification}
\label{sec:random_walk}
Most non-autoregressive TTS systems offer full controllability of the phoneme durations during the inference process. In this work we consider two different approaches to manipulate the phoneme duration. The first approach is to scale each prediction by a constant factor $\bar{\alpha}$ in order to modify the output length before casting to an integer. The second approach modifies the durations individually using a scale per phoneme based on a random walk. Using a random walk approach results in similar modification of neighboring phonemes. In contrast to e.g. drawing fully random durations we expect more realistic duration curves. Given the predicted durations $d_n$ for each of the $N$ input phonemes we compute scaling factors $\alpha_1^N$ as:
\vspace{-0.5em}
\begin{align}
\alpha'_0 = 0.0
&&
\alpha'_{n+1} = \alpha'_n + \mathcal{N}(0, \sigma)
\end{align}
\vspace{-2.0em}
\begin{align}
\alpha_n = 1 + \alpha'_n - \frac{1}{N}\sum_{k=1}^N \alpha'_k && d'_n = d_n \cdot \alpha_n
\end{align}
and clip the values into the range of $[0.9,1.2]$. The predicted durations are then scaled with the respective $\alpha_n$ before applying integer rounding. Although Gaussian upsampling allows for non-rounded durations, we found no significant difference in performance, thus we do not include such experiments in this work. We choose the upper limit larger than the lower limit to have outliers for longer durations which are more prominent in duration distributions from the aligners.  We chose $\sigma$ between 0.0125 and 0.05 as larger values result in excessive clipping. \alert{While there are more sophisticated methods for diverse duration modeling, such as the stochastic duration predictor from VITS \cite{pmlr-v139-kim21f}, our method can be applied to any existing model with duration prediction on-the-fly. No reformulation of the model or additional training is needed.}
\subsection{Metrics}

Unfortunately, evaluating alignments and TTS systems is difficult, as for TTS systems there exists no standardized metric like word-error-rate (WER) for ASR. \alert{Objective metrics such as Mel-Cepstral-Distance (MCD) \cite{407206} or AutoMOS \cite{45744,huang22f_interspeech} are often only valid within the scope of a single publication, as they have trainable or manually adjustable parameters and are thus hard to reproduce.
}
Our final target metric is thus how much we can improve ASR systems by incorporating data synthesized by a TTS system into the ASR training (named \textbf{AED-ASR}). To determine the correctness of the TTS output we report the WER of recognizing the synthesized cross-validation (CV) set with a strong ASR system trained on all of LibriSpeech. The CV set is included in training to reduce errors made by the ASR system itself. To not confuse this with the WER for evaluating an ASR system, we refer to this metric as ``synthetic WER'' (\textbf{sWER}). In addition we propose the mean Kullback-Leiber divergence (\textbf{KLd}) of the durations grouped per ARPA phoneme compared to the duration distribution given by the aligner. \alert{Of course the target distribution of the aligner is only an estimate of the underlying real distribution. Nevertheless, we hypothesize that if the TTS system predicts phoneme durations that follow a more similar distribution this results in ``better'' synthetic data, meaning that the final performance of the ASR system increases.}

\section{Experiments}

\subsection{Data and Training}

We are using the train-clean-100 part of LibriSpeech as annotated training data, and the train-clean-360 part of LibriSpeech as text only data. This is a standard task for many publications on synthetic data generation \cite{Laptev-2020-YouDoNotNeedMore, Baskar-2021-EatEnhancedASR-TT, 9688255}. For the ASR recognition we segment the text using byte-pair encoding \cite{sennrich-etal-2016-neural} with 2000 merge operations and evaluate on LibriSpeech test-clean and test-other. As CV set for the TTS system and for the sWER computation we split 1004 sequences (4 per speaker) from the \textit{train-clean-100} part. The reference WER on the real CV set is 2.0\%. Both the CTC and the HMM-GMM aligner are trained for 100 epochs\footnote{for the HMM-GMM we consider each full EM-step as epoch}, the NAR-TTS for 200, and the ASR for $\sim$165 epochs. In case of the combined training, we over-sample the real data by factor~3 and add the synthetic data once, resulting in a balanced data between real and synthetic data. We then train on the combined data for 55 epochs. All training runs were conducted using a single consumer GPU with 11Gb VRAM (e.g. Nvidia 1080/2080 TI), so the entry barrier to reproduce the results is low. The usual training time for the TTS is 60h, for the ASR 140h and 240h in case of combined training. All models are trained using Adam \cite{DBLP:journals/corr/KingmaB14} and learning rate decay of factor 0.9 based on the CV score.

%
%

\subsection{Effect of vocoding and data preprocessing}


\begin{table}
\caption{WER results for different data conditions. ``Real'' is the original \textit{LS-100h} corpus. ``Vocoder only'' means we extract log-mel from the original data and only apply Griffin\&Lim, ``Synthetic'' refers to using the baseline TTS trained with durations from the HMM-GMM aligner. The AED-ASR test-sets are LibriSpeech test-clean and test-other.}
\label{tab:voc_pp}
\begin{center}
\resizebox{3.2in}{!}{
\begin{tabular}{|l|c||c|c|c|}
\hline
\multirow{3}{*}{\makecell{Data (\textit{LS-100h})}}  & \multirow{3}{*}{\makecell{Silence \\ Pre-processing}} & sWER & \multicolumn{2}{c|}{AED-ASR} \\
\cline{3-5}
& & WER[\%] & \multicolumn{2}{c|}{WER[\%]} \\
\cline{3-5}
& & cv & clean & other \\
\hline
\hline
\multirow{2}{*}{Real} & no & 2.03 & \phantom{0}7.8 & 19.6\\
 & yes & 3.57 & \phantom{0}8.1 & 20.8\\
\cline{1-5}
\multirow{2}{*}{Vocoder only} & no & 2.22 & \phantom{0}8.0 & 19.8\\
 & yes & 3.75 & \phantom{0}8.5 & 20.8 \\
\cline{1-5}
\multirow{2}{*}{Synthetic} & no & 2.00 & 17.6 & 41.8\\
& yes & 2.48 & 15.3 & 39.4\\
\hline
\end{tabular}
}
\end{center}
\end{table}

\begin{table}
\caption{WER and KLd results for the 100h data generated with the TTS using the respective alignment and optional oracle durations.
*The KLd is measured w.r.t. the aligner duration distribution, so it uses a different reference in each case.}
\label{tab:duration_cheat}
\begin{center}
\resizebox{3.2in}{!}{
\begin{tabular}{|l|c||c|c|c|c|}
\hline
\multicolumn{1}{|c|}{\multirow{3}{*}{\makecell{TTS\\Alignment}}}  & \multirow{3}{*}{\makecell{Oracle \\ Durations}} & sWER & KLd & \multicolumn{2}{c|}{AED-ASR} \\
\cline{3-6}
& & WER[\%] & $\Delta$ & \multicolumn{2}{c|}{WER[\%]} \\
\cline{3-6}
& & cv & LS-100 & clean & other \\
\hline
\hline
\multirow{2}{*}{HMM-GMM} & no & 2.48 & 0.069* & 15.3 & 39.4\\
 & yes & 2.96 & 0 & 11.2 & 33.1\\
\cline{1-6}
\multirow{2}{*}{Neural-CTC} & no & 3.37 & 0.085* & 17.2 & 42.4\\
 & yes & 4.53 & 0 & 11.0 & 33.2 \\
\hline
\end{tabular}
}
\end{center}
\end{table}

In \cite{Laptev-2020-YouDoNotNeedMore} it was shown that using a neural vocoder is beneficial over G\&L for creating additional training data. In our case, we wanted to verify how much performance we are loosing by just converting the real data into the log-mel representation and converting it back to audio via G\&L. Table \ref{tab:voc_pp} shows that there is only a small degradation caused by the vocoding process, even if the resulting audio has severe degradation for human listeners due to the incorrectly reconstructed phase information. Another aspect is the silence-pre-processing. In \cite{Rossenbach-2020-GeneratingSynthetic} it was stated that the silence pre-processing is strictly needed to train an autoregressive Tacotron-2 on the ASR data. \alert{It is known that the LibriSpeech corpus has a very unnatural utterance composition, and it often contains very long ($>1s$) portions which are difficult to handle for the TTS system \cite{Zen-2019-LibriTTSACorpusD,Rossenbach-2020-GeneratingSynthetic}}.
As non-autoregressive models are much more robust, we wanted to see how the silence-pre-processing influences the ability to re-synthesize the training data. In Table \ref{tab:voc_pp} a strong degradation is visible for applying the pre-processing on the real data. But while training the NAR-TTS on the non-pre-processed data is stable, the resulting synthetic data yields worse results, so we kept using the silence pre-processing for all further experiments.

\subsection{Effect of oracle durations}


\begin{figure}
\includegraphics[scale=0.72]{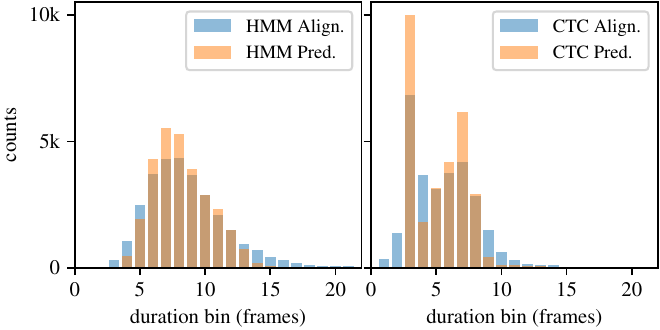}

\caption{Aligned and predicted duration distribution for the ARPA phoneme ``SH'', which depicts typical characteristics for both HMM and CTC based duration distributions. For the prediction we use the TTS trained on the respective alignment.}

\label{fig:durations}
\end{figure}

Table \ref{tab:duration_cheat} shows the results for synthetic data training given the different alignments for the TTS model. It also shows the respective result on synthetic data that was created via providing the given durations from the original alignment. Here it is visible that using the given durations from alignments during synthesis result in better ASR performance. It is also visible that the TTS trained on CTC aligner shows a slightly weaker performance during normal synthesis, but is on par when using the reference alignment. \alert{The rather small gap in performance follows the findings of \cite{Zalkow-2023-EvaluatingSpeechPh}, where it was shown that the choice of the used aligner has a rather low impact on speech synthesis quality.}
When looking at the duration histograms in detail we found that the CTC-alignment-based TTS durations would have a more inconsistent shape. This is also reflected in a higher KLd. As representative for many of the histograms, we show an example in Figure \ref{fig:durations}. While the prediction for the HMM-GMM model is shaped similarly to the alignment, with just a more narrow main peak, the CTC model has an unexpected mismatch for a duration of 3 and 4 frames. This happens for many of the other phonemes as well, and we have no clear explanation for this. Due to the nature of how the alignments are computed, the CTC alignment has generally shorter spoken phonemes and more frames aligned to the ``[space]'' symbol. In contrast, the HMM imposes a minimum duration of 3 frames per phoneme and is more hesitant to align silence.
We also observed that the duration predictor had a higher cross-validation mean-absolute-error loss value when provided with the targets from the CTC aligner. This means the targets were more difficult to learn, which is consistent with the observations made when looking at the histogram plots.

\subsection{Duration Control}

%

\begin{table}
\caption{WER and KLd results for the 100h data generated with the TTS using the HMM-GMM aligner and optional modifications to the phoneme duration prediction during generation.}
\label{tab:duration_mod}
\begin{center}
\resizebox{3.2in}{!}{
\begin{tabular}{|l|l|c||c|c|c|c|}
\hline
\multicolumn{1}{|c|}{\multirow{3}{*}{\makecell{Scaling\\Approach}}}  & \multicolumn{1}{c|}{\multirow{3}{*}{\makecell{Parameter\\Value}}} & \multirow{3}{*}{\makecell{Data\\Length}} & sWER & KLd & \multicolumn{2}{c|}{AED-ASR} \\
\cline{4-7}
& & & WER[\%] & $\Delta$ & \multicolumn{2}{c|}{WER[\%]} \\
\cline{4-7}
& & & cv & LS-100 & clean & other \\
\hline
\hline
\multirow{4}{*}{Constant} & $\alpha = 0.9$ & 73.0 & 3.17 & 0.119 & 16.5 & 40.3\\
 & $\bar{\alpha} = 1.0$ & 81.1 & 2.48 & 0.069 & 15.3 & 39.4\\
 & $\bar{\alpha} = 1.1$ & 89.2 & 2.07 & 0.100 & 15.7 & 39.2\\
 & $\bar{\alpha} = 1.2$ & 97.4 & 1.96 & 0.206 & 16.6 & 41.1\\
\hline
\hline
\multirow{4}{*}{Random walk} & $\sigma = 0.0125$ & 81.3 & 2.62 & 0.059 & 14.5 & 38.0 \\
 & $\sigma = 0.025$ & 82.2 & 2.74 & 0.047 & 13.9 & 36.8 \\
 & $\sigma = 0.0375$ & 82.9 & 2.80 & 0.041 & 13.5 & \textbf{36.4} \\
 & $\sigma = 0.05$ & 83.4 & 2.94 & 0.038 & \textbf{13.4} & 37.2 \\
\hline
\hline
Real Durations & - & 88.3 & 2.96 & 0.0 & 11.2 & 33.1\\
\hline
\end{tabular}
}
\end{center}
\vspace{-1.5em}
\end{table}

As the TTS model was generally predicting about 8-9\% shorter sequences than seen in training, we initially tried to just give a global scaling factor to all durations to cover for this. As shown in Table \ref{tab:duration_mod} this mostly resulted in degradation, although the final data length matched. As we looked at the duration distributions like Figure \ref{fig:durations}, we observed that the majority of predictions was centered in the correct range and the general mean duration shift was mostly caused by long outliers. To ``widen'' the predictions we used our random-walk-based duration modification from Section \ref{sec:random_walk} with different variances $\sigma$. Table \ref{tab:duration_cheat} shows that the more we artificially widen the distributions using the random-walk approach, the smaller the distribution difference is. We also see that the generated training data contributes more, resulting in an improvement of up to 15\% relative in WER of the AED-ASR.

\subsection{Combined Training}
\label{sec:combined_training}

\begin{table}
\caption{WER results for the combined training on LibriSpeech train-clean-100h using the text of train-clean-360h for synthetic data generation. Includes some of the best results from the literature on the exact same task. All results include an external LM during recognition trained on the LibriSpeech LM data.}
\label{tab:combined}
\begin{center}
\resizebox{3.2in}{!}{
\begin{tabular}{|l|l|c|l||c|c|}
\hline
\multirow{3}{*}{Work} & \multicolumn{1}{c|}{\multirow{3}{*}{\makecell{Data}}} & \multicolumn{1}{c|}{\multirow{3}{*}{\makecell{TTS\\ / Aligner}}} & \multirow{3}{*}{Modification} & \multicolumn{2}{c|}{AED-ASR} \\
\cline{5-6}
& & & & \multicolumn{2}{c|}{WER[\%]} \\
\cline{5-6}
& & & & clean & other \\
\hline
\hline
\multirow{8}{*}{Ours} & 100h & - & \multicolumn{1}{c||}{-}  & 6.0 & 15.2\\
\cline{2-6}
 & \multirow{6}{*}{100h + text} &  \multirow{6}{*}{\makecell{NAR-TTS\\ / HMM-GMM}} &  \multicolumn{1}{c||}{-} & 3.6 & 11.8\\
 & & & $\alpha=1.1$ & 3.6 & 12.0 \\
 & & & $\sigma=0.0125$ & 3.5 & \textbf{11.3} \\
 & &  & $\sigma=0.025$ & \textbf{3.4} & 11.4 \\
 & & & $\sigma=0.0375$ & 3.5 & 11.6 \\
 & &  & $\sigma=0.05$ & 3.6 & 11.7 \\
\cline{2-6}
& 460h & - & \multicolumn{1}{c||}{-}  & 3.0 & \phantom{0}8.9\\
\hline
\hline
\multirow{2}{*}{\cite{9688255}} & 100h & - & \multicolumn{1}{c||}{-}  & \phantom{0}5.3 & 14.8 \\
 & 100h + text & Tacotron-2/ - & \multicolumn{1}{c||}{-}  & \phantom{0}3.3 & 12.4 \\
 \hline
 \hline
 \multirow{3}{*}{\cite{Baskar-2021-EatEnhancedASR-TT}} & 100h & - & \multicolumn{1}{c||}{-}  & 14.4 & 36.9 \\
 & 100h + text & Transformer/ -  & \multicolumn{1}{c||}{-}  & \phantom{0}4.7 & 15.2 \\
 & 460h & - & \multicolumn{1}{c||}{-}  & \phantom{0}3.5 & 12.6 \\
 \hline
\end{tabular}
}
\end{center}
\vspace{-1.5em}
\end{table}

We also evaluate how much the ASR system can be improved by synthesizing the text of \textit{train-clean-360} using the TTS models. Table \ref{tab:combined} shows the results for the same ASR model trained on differently generated synthetic data and including an external LM. It can be seen that using synthetic data generated by the TTS improves the ASR performance by over 20\% relative. Using the random-walk modification during synthesis the result improves by another 0.5\% absolute on test-other. Contrary to training on synthetic data only, the performance is better for smaller values of $\sigma$. Nevertheless, we see that there can be a notable improvement on the ASR training when using the random-walk modification. We also show a selection of comparable best results from the literature. With our pipeline we achieve new state-of-the-art performance on the 100h semi-supervised LibriSpeech task.

%
%

\section{Conclusions}

In this work we have shown the effect of temporal variability in the synthetic training data generated by a non-autoregressive TTS. This effect has been studied within a common semi-supervised recognition task. Furthermore, we analyzed the miss-prediction of phoneme durations using oracle setups guiding the TTS generation by durations given from the external aligner computed on real data. By using a random-walk approach we could modify the predicted durations to follow a wider distribution, and improve the quality of the synthetic data w.r.t. to ASR training. With this improvement we achieved new state-of-the-art results by generating synthetic data for the LibriSpeech semi-supervised 100h task.
\vspace{-0.2em}
\section{Acknowledgments}
\vspace{-0.3em}
{This work was partially supported by NeuroSys, which as part of the initiative “Clusters4Future” is funded by the Federal Ministry of Education and Research BMBF (03ZU1106DA), and by the project RESCALE within the program \textit{AI Lighthouse Projects for the Environment, Climate, Nature and Resources} funded by the Federal Ministry for the Environment, Nature Conservation, Nuclear Safety and Consumer Protection (BMUV), funding ID: 67KI32006A.}

\bibliographystyle{IEEEbib}
\bibliography{strings,refs}

\end{document}